\documentclass[letterpaper]{article}
\usepackage{flairs}%aaai
\usepackage{times}
\usepackage{helvet}
\usepackage{courier}

\usepackage{amsmath,amsfonts,amssymb} % enable \mathbb, \mathcal
\usepackage{microtype}      % make text margins prettier
\usepackage{graphicx}       % import graphical images
\usepackage[T1]{fontenc}    % fix font related glitches such as "|"
\usepackage{multirow}       % merge rows in tables
\usepackage{subcaption}     % create sub-tables and sub-figures
\usepackage{bold-extra}     % enable \texttt{\textbf{}}
\usepackage{bm}             % enable bold font in math mode
\usepackage[hang,flushmargin]{footmisc}  % minimize footnote indentation
\usepackage{tabu}
\usepackage{dashrule}
\usepackage{arydshln}
\usepackage{nameref}
\usepackage{url}

\newcommand\MC[2]{\multicolumn{1}{c#1}{\textbf{#2}}}

\newcommand\MR[2]{\multirow{#1}{*}{#2}}
\newcommand\LN{\linebreak\noindent}

\begin{document}
\title{Evaluation of Unsupervised Entity and Event Salience Estimation}
\author{Jiaying Lu \\
  Department of Computer Science \\
  Emory University \\
  Atlanta, GA 30322 USA \\
  \texttt{jiaying.lu@emory.edu} \\
  \And
  Jinho D. Choi \\
  Department of Computer Science \\
  Emory University \\
  Atlanta, GA 30322 USA \\
  \texttt{jinho.choi@emory.edu} \\
}

\maketitle
\begin{abstract}
\begin{quote}
Salience Estimation aims to predict term importance in documents.
Due to few existing human-annotated datasets and the subjective notion of salience, previous studies typically generate pseudo-ground truth for evaluation. However, our investigation reveals that the evaluation protocol proposed by prior work is difficult to replicate, thus leading to few follow-up studies existing. Moreover, the evaluation process is problematic: the entity linking tool used for entity matching is very noisy, while the ignorance of event argument for event evaluation leads to boosted performance. In this work, we propose a light yet practical entity and event salience estimation evaluation protocol, which incorporates the more reliable syntactic dependency parser. Furthermore, we conduct a comprehensive analysis among popular entity and event definition standards, and present our own definition for the Salience Estimation task to reduce noise during the pseudo-ground truth generation process. Furthermore, we construct dependency-based heterogeneous graphs to capture the interactions of entities and events. The empirical results show that both baseline methods and the novel GNN method utilizing the heterogeneous graph consistently outperform the previous SOTA model in all proposed metrics. 
\end{quote}
\end{abstract}

\section{Introduction} \label{sec:intro}

Automatic extraction of prominent information from text has gained significant interest due to the huge volume of articles produced every day. Among various discourse elements, entity and event play a central role to form the skeleton of the story plot. Although entity and event salient estimation is critical for many downstream tasks, we notice that few studies follow-up Xiong et al.~\shortcite{xiong2018towards} and Liu et al.~\shortcite{liu-etal-2018-automatic}. The reasons behind that are the replication crisis of previous approaches and flawed evaluation protocol.

As the notion of salience is subjective and varies from person to person, few human-annotated datasets exist. Hence, previous studies \cite{dunietz-gillick-2014-new,xiong2018towards,liu-etal-2018-automatic} typically generate pseudo-ground truth, following the assumption that an entity or event is considered salient if a summary written by humans includes the reference of entity or event. This practical assumption allows researchers to create large-scale pseudo annotations. 
While some prior work reported state-of-the-art performance on their proposed evaluation metrics, the strict dependency of their pre-processing pipelines hindered the reproducibility. Moreover, the noise introduced by entity linking system used in entity salience task and the ignorance of event argument in event salience task led to debatable scores.

In this paper, we investigate flawed evaluation protocols of previous work and present our own evaluation protocol. The proposed evaluation protocol first defines the mention form and extent of entity and event by combining various popular standards. We then employ a light yet accurate pseudo-ground truth annotation process \footnote{The pseudo annotation is available at \url{https://github.com/lujiaying/pseudo-ee-salience-estimation}} that leverages more reliable dependency parser \cite{he:20a}. Same ranking metrics on top-k salient entities and events are incorporated as evaluation metrics. We argue that the proposed evaluation protocol is easier to replicate by other researchers, and the pseudo ground-truth contains less noise. Moreover, we conduct extensive experiments to re-examine existing approaches and recently proposed Graph Convolutional Neural Networks models that have demonstrated their power in many NLP tasks. Thus, we show that both baseline and novel heterogeneous R-GCN \cite{schlichtkrull2018modeling} methods perform better than the previously proposed models.

\section{Entity and Event (Pseudo) Annotation}
\label{sec:entity-event-annotation}

\begin{table*}[htbp!]
\centering
\begin{subtable}{\textwidth}
\captionsetup{justification=centering}
\centering\small{
\begin{tabular}{c||l|c|c}
\MC{||}{Dataset} & \MC{|}{Typology} & \textbf{Form} & \textbf{Entities} \\
\hline\hline
ACE      & P/O/L/G/F, Vehicle, Weapon + 30 subtypes    & N/NE/NP & 56,898 \\
ECB+     & P/O/L/G/F, Artifact, Generic, Time, Vehicle & N/NE    & 17,187 \\
Rich ERE & P/O/L/G/F + 20 subtypes                     & N/NE/NP & 5,873 \\
RED      & All entity types                            & N/NE    & 10,320 \\
\hline\hline
\bf Ours & All entity types                            & N/NE/NP & 144.3M \\
\end{tabular}}
\caption{Comparisons of entity annotation. Entities: the total number of annotated entities,\\ P/O/L/G/F: Person/Organization/Location/Geopolitical entity/Facility, N/NE/NP: Noun/Named Entity/Noun Phrase.}
\label{tab:name-1}
\end{subtable}
\vspace{0.5em}

\begin{subtable}{\textwidth}
\captionsetup{justification=centering}
\centering\resizebox{\columnwidth}{!}{
\begin{tabular}{c||l|c|c|c}
\bf Dataset & \MC{|}{Typology \bm{$\Rightarrow$} Source} & \textbf{Form} & \multicolumn{2}{c}{\textbf{Events}} \\
\hline\hline
\MR{2}{TimeML}
 & Occurrence, Action, State, Aspectual, Perception, Reporting & \MR{2}{V/N/J/C/PP} & \MR{2}{7,571} & \\                    
 & $\Rightarrow$ 300 docs from news \cite{pustejovsky2003timeml} & & & \\\hline
\MR{2}{ACE}  
 & Business, Conflict, Contact, Justice, Life, Movement, Personnel, Transaction& \MR{2}{V/J} & \MR{2}{5,312} & \MR{2}{\checkmark} \\
 & + 33 subtypes $\Rightarrow$ 599 docs from news and broadcasts \cite{doddington2004automatic} & & \\\hline
\MR{2}{ECB+}
 & Occurrence, State, Aspectual, Causation, Generic, Perception, Reporting & \MR{2}{V/N/J} & \MR{2}{14,884} & \MR{2}{\checkmark} \\
 & $\Rightarrow$ 982 docs from news \cite{cybulska-vossen-2014-using} & & & \\\hline
\MR{3}{Rich ERE}
 & Business, Conflict, Contact, Generic, Irrealis, Justice, Life, Manufacture, & \MR{3}{V/N/J/R} & \MR{3}{2,933} & \MR{3}{\checkmark} \\
 & Movement, Personnel, Transaction + 38 subtypes & & \\
 & $\Rightarrow$ 505 docs from news and discussion forums \cite{song-etal-2015-light} & & \\\hline
\MR{2}{RED}
 & Occurrence, Action, State, Process & \MR{2}{V/N/J} & \MR{2}{8,731} & \\
 & $\Rightarrow$ 95 docs from news and discussion forums \cite{ogorman-etal-2016-richer} & & \\\hline
\hline
\MR{2}{\bf Ours}
 & Occurrence, Action, State, Process & \MR{2}{V/N} & \MR{2}{58.9M} & \MR{2}{\checkmark} \\
 & $\Rightarrow$ 1.65M docs from news and scientific articles & & 
\end{tabular}}
\caption{Comparisons of event annotation. Events: \# of annotated events, \checkmark: events are annotated with their arguments, \\ V/N/J/R/C/PP: predicative Verb/Noun/adJective/adveRb/Clause/Preposition Phrase.}
\label{tab:name-2}
\end{subtable}
\vspace{-0.2cm}
\caption{Comparisons of entity and event annotation in the previous and our works.}
\label{tab:entity-event-definitions}
\end{table*}

Various tasks in entity and event processing have been introduced by previous studies \cite{ruppenhofer-etal-2010-semeval,minard-etal-2015-semeval,moro-navigli-2015-semeval,mitamura:2017,heng:2019}. However, many of them have defined their own terms of entities and events \cite{bies-etal-2016-comparison}, making it difficult to decide which standard to follow. Table~\ref{tab:entity-event-definitions} compares the entity/event annotation from the previous works to the one produced by this work. Our dataset is much larger than the others because it is pseudo-generated. Manual annotation of these kinds requires intensive human labor with a solid linguistics background, which makes the process resource-consuming. Such a large amount of pseudo-annotation in our dataset allows researchers from diverse disciplines to perform big data analysis on entities and events occurring over a long period of time.

The trend is clear that the more recent work it is, the more entity and event types are covered by the work.
Our annotation is the most similar to the RED dataset \cite{ogorman-etal-2016-richer} that covers the broadest types of entities and events, although entities in RED are annotated on the heads unless they are named entities\LN whereas they are annotated on the base Noun Phrases in our dataset, which is more explicit (Section~\ref{ssec:entity-definition}).
Moreover, events are annotated with their core arguments in our dataset, which are not present in RED.
It is worth mentioning that RED comprises extra layers of annotation (e.g., polarity, modality) as well as predicative adjectives for events not included in our dataset, which we will explore in the future.

\subsection{Entity Definition}
\label{ssec:entity-definition}

Inspired by definitions in previous work, we consider all \textbf{base Noun Phrases} (base NPs) as entity candidates, excluding eventive nouns. Although some base NPs, such as \textit{president} or \textit{two weeks}, may refer to multiple real-world objects which violate the rigid definition, this simplification is beneficial to pseudo annotation. Another advantage of using base NPs instead of higher-level NPs is the more fine-grained entity annotation.

\subsection{Event Definition}
\label{ssec:event-definition}

An event describes ``\textit{who did what to whom when and where}''. Therefore, the event trigger (\textit{what}) itself can not represent a complete event. In our definition, event triggers and core arguments are essential components of events. \textbf{Verbs}, \textbf{Eventive Nouns} (deverbal nouns, proper names referring to historically significant events), phrase constituted by \textbf{Light Verb + Noun} and predicative \textbf{\textit{Adjectives}} are generally considered as \textbf{event triggers} in prior studies. For annotation simplicity, adjectives are not included in our definition. Since not all verbs and nouns are valid event triggers, we create a pre-defined vocabulary using following procedures: 

\begin{enumerate}
    \item We collect the candidate list of verbs and deverbal nouns from FrameNet \cite{baker1998berkeley} and NomBank \cite{meyers-etal-2004-nombank} utilizing the same 569 frames generated by \cite{liu-etal-2018-automatic}. 
    \item We then manually add valid head words (around 47) of proper names such as \textit{epidemic, earthquake}, etc.
    \item Similar to previous work, we remove auxiliary and copular verbs\footnote{Auxiliary and copular verbs include \textit{appear, become, do, have, seem, be}.}, light verbs\footnote{Light verbs include \textit{do, get, give, go, have, keep, make, put, set, take}.}, and report verbs\footnote{Report verbs include \textit{argue, claim, say, suggest, tell}.}, as they are rarely representative events.
\end{enumerate}

\noindent This gives us a total of 2645 verb lemmas and 516 eventive nouns. Regarding core arguments, we consider entity and sub-event participants specifying \textit{who} or \textit{whom} is involved in the event.  

\subsection{Pseudo Annotation}
\label{ssec:pseudo-annotation}

For articles in the Corpus, we annotate pseudo salient entities and events by extracting entity, event trigger, event core argument mentions in the human written abstracts according to the aforementioned definitions. Dependency parser is the main tool we leverage to conduct the pseudo annotation. We believe the dependency parser is more reliable than the entity linker used in previous studies, since the SOTA model achieved above 97\% Unlabeled Attachment Score (UAS) on Penn Treebank \cite{he:20a} and the parser is easy to integrate with other NLP tools. The pseudo annotation process is as follows: First, sentences in the abstract are parsed by a syntactic dependency parser. Second, heuristics\footnote{heuristics for base NP using context-free grammar: \texttt{NP->DT $\bar{N}$; $\bar{N}$->NN; $\bar{N}$->NN $\bar{N}$; $\bar{N}$->JJ $\bar{N}$; $\bar{N}$->$\bar{N}$ $\bar{N}$}} are applied to identify \textit{base NPs} and \textit{verbs} upon POS taggers and dependency relations \cite{DBLP:conf/tlt/Choi17}. Then, all base NPs excluding ones that have eventive noun heads are extracted as entity candidates, while complement base NPs are event candidates. Finally, verbs or the action carrier verbs within verb phrases (i.e. discarding auxiliary verbs and/or raising verbs in verb phrases) in the pre-defined vocabulary are extracted as event trigger candidates. In the above eventive noun and verb cases, only the head part would be annotated: 

\begin{center}
\textit{Allen was \underline{walking} quickly to the mall.}\\
\textit{This terrible \underline{war} could have \underline{ended} in a month.}
\end{center}

\noindent Besides, in the \textit{light verbs + noun} structure, only the nominal part (typically eventive noun) would be annotated:

\begin{center}
\textit{John made a \underline{call} to Mary.}\\
\textit{John gave an \underline{interview} to Mary.}
\end{center}

\noindent For event core arguments, we consider entities or sub-events that have \textit{nominal subject}, \textit{object} or \textit{dative} relations with event trigger. For instance, \textit{John} and \textit{Mary} are annotated as core arguments in above cases.

\section{Proposed Approach} \label{sec:approach}

The salience identifying ability of humankind depends significantly on relational structures of information. Intuitively, we construct an undirected graph per input article which distills the important knowledge for entities and events salience estimation. The graph $\mathcal{G}$ is a heterogeneous graph that consists of two types of nodes: entity nodes, and event nodes. The node types can be further expanded by including coreferential cluster nodes, sentence nodes, and others when necessary. The details about node candidate generation is described in subsection \nameref{ssec:candidate_generation}. 
%The candidate generation module parses every sentence in the document $\mathcal{D}$ into the dependency tree and extracts \textit{entity spans} $\mathcal{X}_{entity}$, and \textit{event spans} $\mathcal{X}_{event}$ according to the POS taggers and syntactic relations. After all candidates are extracted, the graph $\mathcal{G}$ is then expanded with these spans as new nodes interchangeably denoted as $\mathcal{V}$. 
Then the graph construction module collects these generated entity and event nodes and connects them accordingly. Different from most previous work which considers the graph as a fully-connected graph, edges are added according to the dependency tree arcs which lead the plot graph $\mathcal{G}$ a partially-connected graph. Multiple types of edges including dependency edges, coreferential edges, and inter-sentence edges are added to better reflect semantic and syntactic relations between nodes.

%The dependency-based edges can better reflect relations between nodes compared to previous fully-connected edges. In addition, more types of edges are added to $\mathcal{G}$ to comprehensively capture information required to determine the salient nodes.  For main entity and event of news article setting, coreferential cluster edges are incorporated to handle the abundant expressions of the same object. Sentence edges are included to address the lack of inter-sentence level information flow for intra-sentence level dependency trees.

Once the nodes and edges are established, edge weights are critical for salience node prediction. For initial edge weights, we design multiple mechanisms to utilize the frequency, position, and other information as much as possible. The weighted graph can be directly used for classic graph ranking algorithms. However, heuristic-based weights design can be tricky and hard to generalize. To overcome the disadvantages above, we train a heterogeneous R-GCN \cite{schlichtkrull2018modeling} model to learn the node salience by feeding top frequent entities and events as the alternative training data.

\subsection{Candidate Generation Module} \label{ssec:candidate_generation}

\begin{figure}[!htbp]
\centering
  \includegraphics[width=\columnwidth]{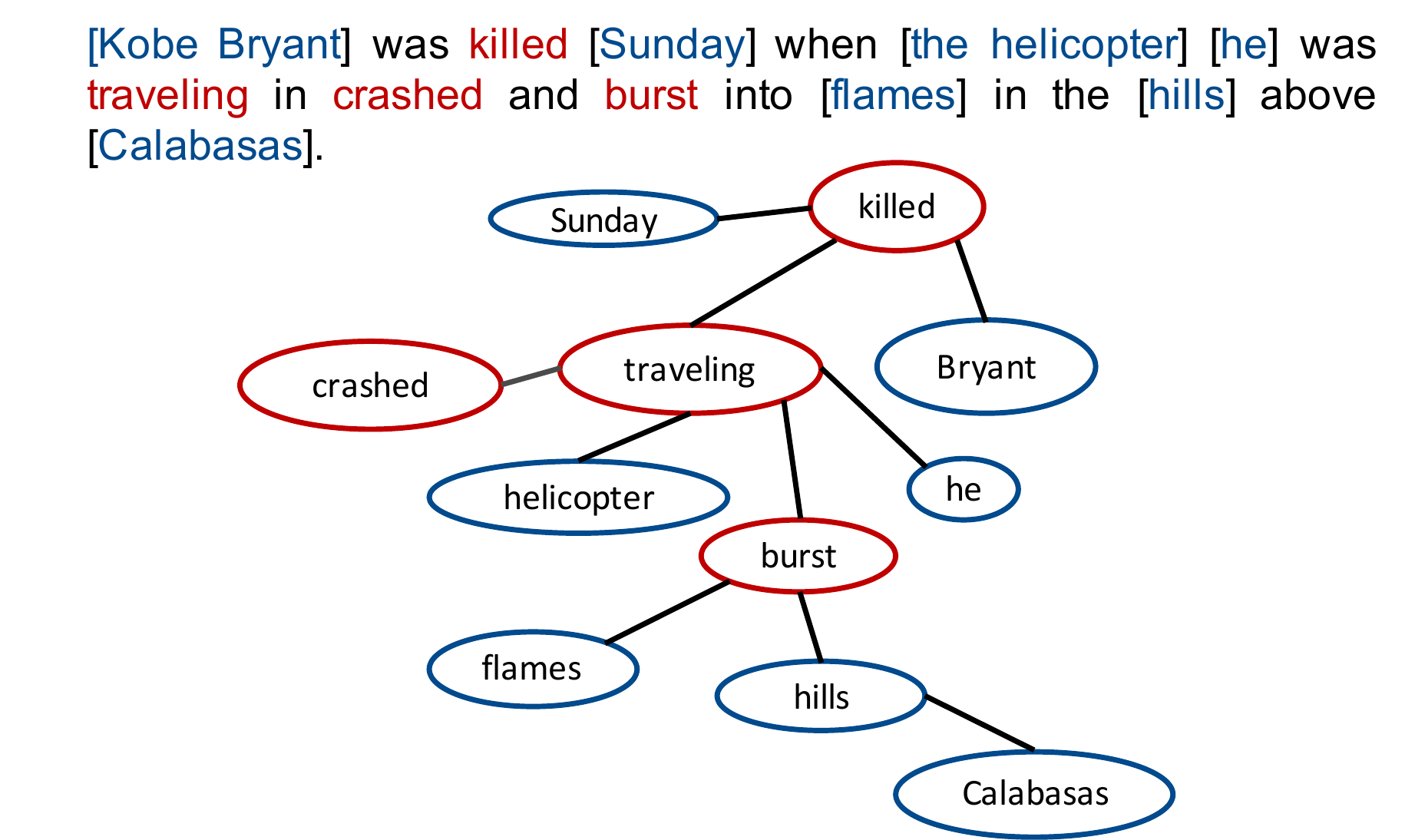}
  \caption{Dependency-based Candidate Generation}
  \label{fig:span_extraction}
\end{figure}

\noindent Figure \ref{fig:span_extraction} shows an example of entity and event candidate spans generated by a dependency tree. For simplicity, each span in figure \ref{fig:span_extraction} is represented by the dependent head. The extraction result can be regarded as the remaining dependency tree after removing auxiliary words. 

As mentioned in Section \ref{sec:entity-event-annotation}, entities and events have their own syntactic properties:

1) Entities are generally \textit{noun phrases}.

2) Events are \textit{verbal predicates} or \textit{nominal predicates}.
\vspace{1.0ex}

\noindent It is then intuitively to derive context-free grammar rules which leverage the POS tagging and dependency relation information to extract entities and events. Since the extraction process is relatively deterministic, the result is reliable. Our implementation uses a BERT-based Dependency Parser from \citeauthor{he:20a} \shortcite{he:20a} and can be extended using any dependency parser.

\subsection{Graph Construction Module}

After all entity and event candidates are extracted, it is natural to construct a graph to capture the relation structure of information. In addition to the intra-sentence dependency relations from dependency parsing, inter-sentence level relations such as adjacent sentence syntactic roots, coreference resolution is also introduced in our graph construction model. The graph ranking algorithm is then deployed for salience identifying. 
In the following subsections, we will describe the proposed construction algorithm for the graph.

\subsubsection{Dependency-based Graph Construction} \label{sssec:dep_tree_graph}

\begin{figure}[!htbp]
\centering
  \includegraphics[width=\columnwidth]{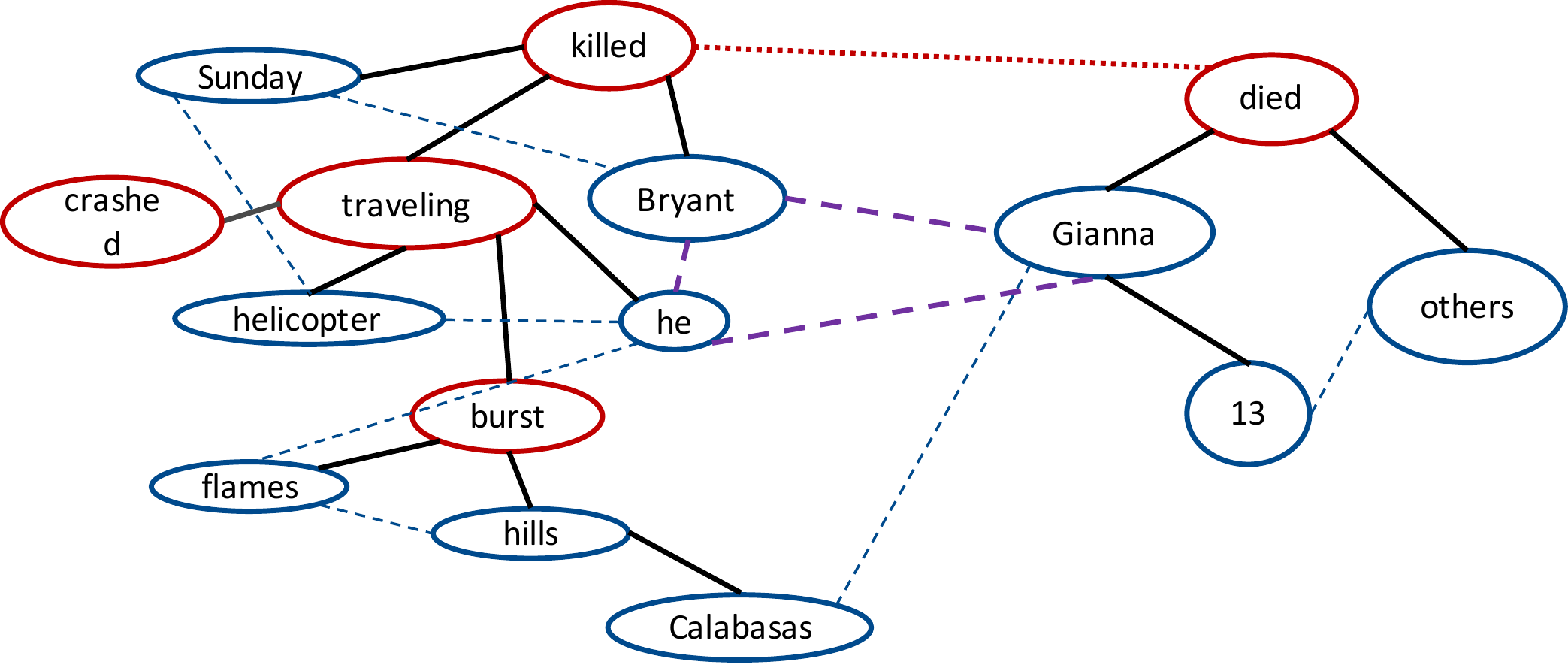}
  \caption{Dependency-based Graph with extra edges introduced. Dependency, Coreferential, and Adjacent sentence edges are denoted by blue, purple and red edges, respectively.}
  \label{fig:dep_tree_graph}
\end{figure}

Figure \ref{fig:dep_tree_graph} shows a dependency-based graph for two sentences about Kobe Bryant's helicopter crash. Adjacent noun phrases edges, coreferential edges, and adjacent sentence syntactic roots edges are further included in the graph to capture more information.
First, the initial dependency-based graph is already established after the candidate extraction phase. The edge $w_{ij}^{dep}$ between two nodes $v_i$ and $v_j$ are weighted according to the sum of the inverse tree distance between each span headword of $hw_i$ and $hw_j$, as equation \ref{eq:tree_dis_weight} denoted. The tree distance is defined as the minimum number of edges to be traversed to reach one node from another. It is also worth noting that in dependency-based graph nodes that share the same span have not been merged yet. 

\begin{equation} \label{eq:tree_dis_weight}
\small
    w_{ij}^{dep} = \frac{1}{tree\_dist(hw_i, hw_j)},\ \forall e_{ij} \in E_{dep}
\end{equation}

\noindent Next, adjacent NPs and syntactic roots are connected accordingly. Edges between Adjacent NPs($E_{NPs}$) reflect the information propagation between neighbor entities and contain both intra- and inter-sentence level information. The edge weights $w^{NPs}$ are set by the inverse of tree distances as well. On the other hand, edges between adjacent syntactic roots $E_{roots}$ reflect the information propagation between neighbor sentences. Since the dependency tree only provides within sentence distances, we can assign a constant to $w^{roots}$ (in practice, we choose $1$) as the virtual distance for each edge of $E_{roots}$. Therefore, equation \ref{eq:tree_dis_weight} can be expanded to include edges in $E_{NPS}$ and $E_{roots}$.

Moreover, spans in same coreferential clusters are connected. The coreferential clusters can be obtained from either ground truth of the corpus or system-generated result. The edge weights $w^{coref}$ of $E_{coref}$ can also be assigned to a constant (we still use $1$ in experiments) because spans in one cluster are equally important. 
Finally, NP spans consisting of the same tokens within one article $\mathcal{D}$ would then be merged into one node $x$ in the plot graph $\mathcal{G}$, except single pronoun spans. Therefore,

\begin{equation} \label{eq:span_merge}
x_i = \{s_{i,j}, ..., s_{i,k}\}\ where\ s_{i,j} = s_{i,k}
\end{equation}

\noindent In equation \ref{eq:span_merge}, $s_{i,j}$ denotes the \textit{jth span} of node $x_i$. 
The overall weights between node $x_i$ and $x_j$ then become the sum of different types of weights, which is shown in Equation \ref{eq:dep_based_weight_sum}. $w^{dep}_{kl}, w^{NPs}_{kl}, w^{roots}_{kl}, w^{coref}_{kl}$ represent edge weights from dependency tree, adjacent Nps, adjacent syntactic roots and corference resolution clusters between $span_{i,k}$ and $span_{j,l}$, respectively. 

\begin{equation} \label{eq:dep_based_weight_sum}
\small
    w_{ij}=\sum_{s_{i,k} \in x_i} \sum_{s_{j,l} \in x_j} \frac{1}{w^{dep}_{kl}+w^{NPs}_{kl}+w^{roots}_{kl}+w^{coref}_{kl}}
\end{equation}

\noindent After the edge weights are set, classic unsupervised graph ranking such as TextRank \cite{mihalcea-tarau-2004-textrank} algorithms can be directly employed. The importance of nodes in the graph is calculated and then used as indicators of whether entities or nodes are salient.

\subsubsection{Graph Neural Network upon Heterogeneous Graph}

The advantage of classic graph ranking methods is that no training data is required. However, the design of edge weights is tricky and may significantly impact the final performance. A learning-based model is preferable to let the model automatically learn the importance of different edge types given source and target vertices. Therefore, a relational GCN \cite{schlichtkrull2018modeling} is adapted. The key idea of relation GCN is that edge weights are determined by edge types and can be learned through a training process. Equation \ref{eq:R-GCN} illustrates the vertex representation update in R-GCN, where $c_{i,r}$ denotes a normalization constant and $W_r$ denotes the learnable weight for edge type $r$. The choice of initial node embedding is flexible. Bag-of-words embedding, static word embedding, or contextual word embedding can be used, with POS tagging, named entity results as optional extra features. According to the graph construction, edge types include dependency relations, adjacent noun phrase, adjacent sentence root, and coreferent vertices. Since the constructed graph is a heterogeneous graph, the edge types are further categorized by the source and target vertex types.

\begin{equation} \label{eq:R-GCN}
    h_i^{l+1}=\sigma(W_0^lh_i^l+\sum_r \sum_j \frac{1}{c_{i,r}}W_r^lh_j^l)
\end{equation}

\noindent Although no ground-truth entities and events are provided in datasets, we choose top-frequent entities and events as alternative training samples. Thus, the R-GCN model is trained in a supervised node classification setting. During inference, heterogeneous graphs are constructed for input text using the same methods and then used to predict salient nodes.

\section{Experiments} \label{sec:experiments}

\subsection{Datasets}

\begin{table}[htbp!]
\resizebox{\columnwidth}{!}{
\centering
\begin{tabular}{l||l|l|l}
Dataset           & \# of Docs in Trn /Dev /Tst & \# of E\&E & avg of E\&E \\
\hline \hline
NYT    & 526K /64K /64K      & 98M /39M              & 150 /60              \\ \hline
SS & 800K /100K /100K    & 46M /19M                & 46 /20                    \\
\end{tabular}
}
\vspace{-0.2cm}
\caption{Statistics for New York Times and Semantic Scholar datasets. (\small Trn, Dev, Tst: Train, Development, Test set. E\&E: Entity and Event.)}
\label{tab:dataset_stat}
\end{table}

\noindent We evaluate the proposed protocol on a news corpus and a scientific publication corpus as follows.
\begin{itemize}
    \item \textbf{New York Times} \cite{dunietz-gillick-2014-new} provides \textit{654K} online news articles that contains experts written summarizes. The entities and events annotated on the summarizes are considered as salient. 
    \item \textbf{Semantic Scholar} \cite{xiong2017explicit} contains \textit{one million} randomly sampled scientific publications. Since only the abstracts and titles of papers are available, we consider entities that occurred on titles as salient. 
\end{itemize}

\begin{table*}[htb!]
\captionsetup{justification=centering}
\centering \small
\begin{tabular}{l||llll|llll}
& \multicolumn{4}{c|}{\textbf{New York Times}}  & \multicolumn{4}{c}{\textbf{Semantic Scholar}} \\ \hline \hline
\textbf{Method} & \textbf{F1@1} & \textbf{F1@3} & \textbf{F1@5} & \textbf{F1@10} & \textbf{F1@1} & \textbf{F1@3} & \textbf{F1@5} & \textbf{F1@10} \\ \hline 
TextRank  & 3.69 & 8.22 & 10.60 & 13.17 & 5.46 & 8.60 & 9.43 & 9.49\\
KCE       & 5.46 & 11.64 & 14.52 & 16.57 & 4.83 & 7.82 & 8.41 & 7.77 \\ \hline
Frequency & \textbf{7.02} & 14.00 & 17.69 & 22.40 & \textbf{10.78} & \textbf{16.49} & \textbf{17.11} & \textbf{15.08} \\
Location  & 3.94 & 10.26 & 16.40 & \textbf{23.86} & 4.65 & 12.08 & 13.76 & 13.45 \\ \hline
R-GCN       & 6.65 & \textbf{14.03} & \textbf{18.10} & 22.72 & 9.66 &  15.39 &  16.35 & 14.81 \\
\end{tabular}
\vspace{-0.2cm}
\caption{Entity salience estimation F1 scores (in percentage).}
\label{tab:exp_results_entity_own_metrics}
\end{table*}

\subsection{Proposed Evaluation Metrics}

In the paper which \citeauthor{dunietz-gillick-2014-new} \shortcite{dunietz-gillick-2014-new} first introduced precision, recall and $F1$ as evaluation metrics since the task was formulated as a binary classification problem.
In the follow-up studies, \citeauthor{xiong2018towards} \shortcite{xiong2018towards} and \citeauthor{liu-etal-2018-automatic} \shortcite{liu-etal-2018-automatic} argued that the importance of entities and events was on a continuous scale. They then proposed evaluating the salience task by ranking metrics: precision, recall at top 1, 5 and 10 salient entities and events. Although the ranking metrics can reveal the ability of salience identification, the way they calculated metrics was doubtful. For instance, if one mention $x_i=\{t_a, t_b, ... t_z\}$ occurred in the expert-written abstract was accidentally linked to an entity $E_a$ and the mention with exactly same tokens $x_j=\{t_a, t_b, ... t_z\}$ in expert-written abstract was linked to another entity $E_b$, $x_i$ would be considered as non-salient. This violated the assumption that "an entity is salient if a reference summary includes it." 

\noindent \textbf{(Macro-averaged) P, R, F1 @Top-k} To address the disadvantages and the inconsistency of previous evaluations, we propose different metrics for entities and events to better measure the salience estimation ability. Both metrics neglect punctuation marks and articles (\textit{a, an, the}). These metrics highly rely on the pseudo annotated salient entities and events. For entity salience estimation, duplicated mentions of pseudo annotated salient entities and candidate salient entities are first merged separately. Then, precision (P), recall (R), F1 score are calculated between pseudo ground-truth and prediction using \textit{exact match}. %Although exact match may be an overly strict criterion that considers a predicted mention incorrect even if it is a lexical variant of ground-truth, exact match eliminates the noise that is hard to avoid when using automatic tool to identify such name variants. 
For event salience estimation, P, R, F1 are conducted between pseudo annotated event triggers and predicted triggers; meanwhile, between pseudo annotated event triggers with corresponding event arguments and predicted triggers with corresponding arguments. 

In conclusion, our proposed evaluation protocol is superior to previous protocols for following reasons.

1) the flexibility and reproducibility for using any NLP tools to extract entity and event mentions.

2) the finer definition of a complete event that not only considers event trigger but also event arguments.

3) the less noisy dependency parser for pseudo ground-truth annotation.

\begin{table*}[htbp!]
\captionsetup{justification=centering}
\centering
\resizebox{\textwidth}{!}{%
\begin{tabular}{l||llllllll|llllllll}
                      & \multicolumn{8}{c|}{\textbf{New York Times}}  &\multicolumn{8}{c}{\textbf{Semantic Scholar}} \\ 
\hline \hline
\multicolumn{1}{c||}{\multirow{2}{*}{\textbf{Method}}} &
  \multicolumn{2}{c}{\textbf{F1@1}} &
  \multicolumn{2}{c}{\textbf{F1@3}} &
  \multicolumn{2}{c}{\textbf{F1@5}} &
  \multicolumn{2}{c|}{\textbf{F1@10}}&
  \multicolumn{2}{c}{\textbf{F1@1}} &
  \multicolumn{2}{c}{\textbf{F1@3}} &
  \multicolumn{2}{c}{\textbf{F1@5}} &
  \multicolumn{2}{c}{\textbf{F1@10}} \\
\multicolumn{1}{c||}{} & Trig   & +Arg   & Trig   & +Arg   & Trig   & +Arg   & Trig   & +Arg   & Trig   & +Arg   & Trig   & +Arg   & Trig   & +Arg   & Trig   & +Arg\\ \hline
TextRank              & 13.72 & 10.82 & 21.16 & 16.01 & 22.84 & 16.96 & 22.85 & 16.44 & 19.34 & 17.52 & 19.49 & 16.48 & 17.12 & 13.83 & 13.72 & 10.70\\
KCE                   & 13.11 & N/A    & 15.42 & N/A    & 15.65 & N/A    & 15.65 & N/A  & N/A  \\ \hline
Frequency             & \textbf{17.19} & \textbf{12.84} & 24.85 & 18.81 & 26.19 & 19.82 & 24.88 & 18.71 & \textbf{23.92} & \textbf{21.53} & \textbf{23.08} & \textbf{19.73} & \textbf{19.29} & 16.06 & 14.53 & 11.75\\
Location              & 15.36 & 9.82 & \textbf{26.79} & 18.35 & \textbf{30.58} & \textbf{21.44} & 30.45 & \textbf{21.70} & 16.61 & 15.02 & 19.33 & 16.43 & 17.91 & 14.73 & 14.62 & 11.66\\ \hline
R-GCN                   & 16.15 & 12.08 & 25.59 & \textbf{18.98} & 28.12 & 20.91 & \textbf{28.47} & 20.92 & 20.79 & 19.28 & 21.57 & 18.88 & 19.19 & \textbf{16.15} & \textbf{15.00} & \textbf{12.06}\\ 
\end{tabular}%
}
\vspace{-0.2cm}
\caption{Event salience estimation F1 scores (in percentage). \textit{Trig}, \textit{+Arg} denotes evaluation on trigger only and trigger with corresponding arguments, respectively. (\small Event detection was not conducted by KCE on Semantic Scholar as authors stated.)}
\label{tab:exp_results_event_own_metrics}
\end{table*}

\subsection{Compared Methods}

\textbf{Previous Methods} In the experiments, the classic unsupervised method TextRank \cite{mihalcea-tarau-2004-textrank} and the previous SOTA supervised method Kernel-based Centrality Estimation (KCE) \cite{liu-etal-2018-automatic} are analyzed, along with multiple baseline models and our novel unsupervised GNN based model. It is worth noting that KCE only considered event triggers while producing salient events, and KCE did not conduct event salience estimation on Semantic Scholar. Therefore, numbers for corresponding columns in the performance table are blank.

\noindent \textbf{Baseline Methods} For both Frequency and Location methods, we first generate candidate entities and events by employing the same pipelines of heterogeneous graph construction. The frequency method then estimates the salience of an entity by its lemma frequency, and the salience of an event by its frequency of full lemma consisting of trigger and arguments. If two candidates have same frequencies, the one that occurred earlier would be ranked higher. On the other hand, the Location method essentially estimates the term salience by its occurrence position in the text, the earlier the higher.

\noindent \textbf{Proposed Heterogeneous GCN model} We construct a graph per article with 2 node types (entity, event), 92 edge types (bidirectional; including self edge, adjacent node edge, adjacent sentence edge, coreference edge, syntactic dependency edge, same headword edge). Current initial node embedding of is Glove embedding \shortcite{pennington2014glove} concatenating random initialized pos-tagging embedding, and then adding sinusoidal positional embedding. A heterogeneous R-GCN that learns different edge weights for each unique (source node, edge, target node) is applied to capture the characteristics of each node/edge type and to model the interaction between nodes. We treat the top-10 first occurred entities and events as salient terms. Therefore, the heterogeneous R-GCN is trained in the unsupervised setting since no ground-truth data is provided. 

\subsection{Result Analysis}

Table \ref{tab:exp_results_entity_own_metrics} shows the entity salience estimation performance, while Table \ref{tab:exp_results_event_own_metrics} shows the event salience estimation performance on trigger only and trigger with arguments. 
On \textbf{entity salience estimation}, we observe that the Frequency baseline method and our novel heterogeneous R-GCN approach outperforms classic unsupervised TextRank and SOTA supervised KCE significantly for all \textit{F1@1,3,5,10} metrics in both New York Times and Semantic Scholar datasets. On the other hand, the Location baseline method has better performance for all metrics except \textit{F1@1}. Although the proposed R-GCN approach does not achieve the best performance in all metrics, the improvement over SOTA methods is consistent ($1.19\%$ to $6.55\%$ in NYT, $4.83\%$ to $7.94\%$ in SS). It is also worth noting that R-GCN has not achieved the most competitive performance for every metric in Semantic Scholar dataset because titles are considered as the evaluation reference in this setting and titles are generally short sentences containing words occurring repeatedly in documents. 
On \textbf{event salience estimation}, we observe similar performance margins for approaches upon the constructed graph, indicating the effectiveness in leveraging the distilled structured relations as information to identify salient entities and events. Among the approaches utilizing the constructed graphs, baseline methods have advantages in detecting small numbers of top salient terms, while the R-GCN model is an expert at capturing bigger numbers of top salient terms.

\section{Conclusion}
\label{sec:conclusion}

In this paper, we revisit the evaluation metrics of the salience task, and reveal the drawbacks of previous work. To address the experiment reproducibility, noise from entity linker, and duplicated event trigger issues, we propose a light yet practical evaluation protocol that relies on the stable syntactic dependency parser. The empirical experiment result shows that our approaches are better than the SOTA approaches in the proposed metrics, which indicates the effectiveness of the constructed heterogeneous graph. Furthermore, we provide quantitative results on the complete event rather than the mere event trigger. In future work, we will improve our approach by employing an EM-based unsupervised algorithm, aiming to surpass the baseline approaches.

\bibliographystyle{flairs}
\bibliography{flairs}

\begin{thebibliography}{}

\bibitem[\protect\citeauthoryear{Baker, Fillmore, and
  Lowe}{1998}]{baker1998berkeley}
Baker, C.~F.; Fillmore, C.~J.; and Lowe, J.~B.
\newblock 1998.
\newblock The berkeley framenet project.
\newblock In {\em COLING'1998}.

\bibitem[\protect\citeauthoryear{Bies \bgroup et al\mbox.\egroup
  }{2016}]{bies-etal-2016-comparison}
Bies, A.; Song, Z.; Getman, J.; Ellis, J.; Mott, J.; Strassel, S.; Palmer, M.;
  Mitamura, T.; Freedman, M.; Ji, H.; and O{'}Gorman, T.
\newblock 2016.
\newblock A comparison of event representations in {DEFT}.
\newblock In {\em Proceedings of the Fourth Workshop on Events}.

\bibitem[\protect\citeauthoryear{Choi}{2017}]{DBLP:conf/tlt/Choi17}
Choi, J.
\newblock 2017.
\newblock Deep dependency graph conversion in english.
\newblock In {\em Proceedings of the 15th International Workshop on Treebanks
  and Linguistic Theories}.

\bibitem[\protect\citeauthoryear{Cybulska and
  Vossen}{2014}]{cybulska-vossen-2014-using}
Cybulska, A., and Vossen, P.
\newblock 2014.
\newblock Using a sledgehammer to crack a nut? lexical diversity and event
  coreference resolution.
\newblock In {\em LREC'14}.

\bibitem[\protect\citeauthoryear{Doddington \bgroup et al\mbox.\egroup
  }{2004}]{doddington2004automatic}
Doddington, G.~R.; Mitchell, A.; Przybocki, M.~A.; Ramshaw, L.~A.; Strassel,
  S.~M.; and Weischedel, R.~M.
\newblock 2004.
\newblock The automatic content extraction (ace) program-tasks, data, and
  evaluation.
\newblock In {\em LREC}.

\bibitem[\protect\citeauthoryear{Dunietz and
  Gillick}{2014}]{dunietz-gillick-2014-new}
Dunietz, J., and Gillick, D.
\newblock 2014.
\newblock A new entity salience task with millions of training examples.
\newblock In {\em EACL'14}.

\bibitem[\protect\citeauthoryear{He and Choi}{2020}]{he:20a}
He, H., and Choi, J.~D.
\newblock 2020.
\newblock {Establishing Strong Baselines for the New Decade: Sequence Tagging,
  Syntactic and Semantic Parsing with BERT}.
\newblock In {\em FLAIRS'20}.

\bibitem[\protect\citeauthoryear{Ji \bgroup et al\mbox.\egroup
  }{2019}]{heng:2019}
Ji, H.; Sil, A.; Dang, H.~T.; Soboroff, I.; Nothman, J.; and Hub, S.~I.
\newblock 2019.
\newblock {Overview of TAC-KBP 2019 Fine-grained Entity Extraction}.
\newblock In {\em TAC:KBP'19}.

\bibitem[\protect\citeauthoryear{Liu \bgroup et al\mbox.\egroup
  }{2018}]{liu-etal-2018-automatic}
Liu, Z.; Xiong, C.; Mitamura, T.; and Hovy, E.
\newblock 2018.
\newblock Automatic event salience identification.
\newblock In {\em EMNLP'18}.

\bibitem[\protect\citeauthoryear{Meyers \bgroup et al\mbox.\egroup
  }{2004}]{meyers-etal-2004-nombank}
Meyers, A.; Reeves, R.; Macleod, C.; Szekely, R.; Zielinska, V.; Young, B.; and
  Grishman, R.
\newblock 2004.
\newblock The {N}om{B}ank project: An interim report.
\newblock In {\em {HLT}-{NAACL} 2004}.

\bibitem[\protect\citeauthoryear{Mihalcea and
  Tarau}{2004}]{mihalcea-tarau-2004-textrank}
Mihalcea, R., and Tarau, P.
\newblock 2004.
\newblock {T}ext{R}ank: Bringing order into text.
\newblock In {\em EMNLP'04}.

\bibitem[\protect\citeauthoryear{Minard \bgroup et al\mbox.\egroup
  }{2015}]{minard-etal-2015-semeval}
Minard, A.-L.; Speranza, M.; Agirre, E.; Aldabe, I.; van Erp, M.; Magnini, B.;
  Rigau, G.; and Urizar, R.
\newblock 2015.
\newblock {T}ime{L}ine: Cross-document event ordering.
\newblock In {\em {S}em{E}val 2015}.

\bibitem[\protect\citeauthoryear{Mitamura, Liu, and Hovy}{2019}]{mitamura:2017}
Mitamura, T.; Liu, Z.; and Hovy, E.
\newblock 2019.
\newblock {Events Detection, Coreference and Sequencing: What's next?}
\newblock In {\em TAC:KBP'17}.

\bibitem[\protect\citeauthoryear{Moro and
  Navigli}{2015}]{moro-navigli-2015-semeval}
Moro, A., and Navigli, R.
\newblock 2015.
\newblock {S}em{E}val-2015 task 13: Multilingual all-words sense disambiguation
  and entity linking.
\newblock In {\em {S}em{E}val 2015}.

\bibitem[\protect\citeauthoryear{O{'}Gorman, Wright-Bettner, and
  Palmer}{2016}]{ogorman-etal-2016-richer}
O{'}Gorman, T.; Wright-Bettner, K.; and Palmer, M.
\newblock 2016.
\newblock Richer event description: Integrating event coreference with
  temporal, causal and bridging annotation.
\newblock In {\em Proceedings of the 2nd Workshop on Computing News
  Storylines}.

\bibitem[\protect\citeauthoryear{Pennington, Socher, and
  Manning}{2014}]{pennington2014glove}
Pennington, J.; Socher, R.; and Manning, C.~D.
\newblock 2014.
\newblock Glove: Global vectors for word representation.
\newblock In {\em EMNLP'14}.

\bibitem[\protect\citeauthoryear{Pustejovsky \bgroup et al\mbox.\egroup
  }{2003}]{pustejovsky2003timeml}
Pustejovsky, J.; Castano, J.~M.; Ingria, R.; Sauri, R.; Gaizauskas, R.~J.;
  Setzer, A.; Katz, G.; and Radev, D.~R.
\newblock 2003.
\newblock Timeml: Robust specification of event and temporal expressions in
  text.
\newblock {\em New directions in question answering}.

\bibitem[\protect\citeauthoryear{Ruppenhofer \bgroup et al\mbox.\egroup
  }{2010}]{ruppenhofer-etal-2010-semeval}
Ruppenhofer, J.; Sporleder, C.; Morante, R.; Baker, C.; and Palmer, M.
\newblock 2010.
\newblock {S}em{E}val-2010 task 10: Linking events and their participants in
  discourse.
\newblock In {\em Proceedings of the 5th International Workshop on Semantic
  Evaluation}.

\bibitem[\protect\citeauthoryear{Schlichtkrull \bgroup et al\mbox.\egroup
  }{2018}]{schlichtkrull2018modeling}
Schlichtkrull, M.; Kipf, T.~N.; Bloem, P.; Van Den~Berg, R.; Titov, I.; and
  Welling, M.
\newblock 2018.
\newblock Modeling relational data with graph convolutional networks.
\newblock In {\em ESWC}.

\bibitem[\protect\citeauthoryear{Song \bgroup et al\mbox.\egroup
  }{2015}]{song-etal-2015-light}
Song, Z.; Bies, A.; Strassel, S.; Riese, T.; Mott, J.; Ellis, J.; Wright, J.;
  Kulick, S.; Ryant, N.; and Ma, X.
\newblock 2015.
\newblock From light to rich {ERE}: Annotation of entities, relations, and
  events.
\newblock In {\em Proceedings of the The 3rd Workshop on {EVENTS}}.

\bibitem[\protect\citeauthoryear{Xiong \bgroup et al\mbox.\egroup
  }{2018}]{xiong2018towards}
Xiong, C.; Liu, Z.; Callan, J.; and Liu, T.-Y.
\newblock 2018.
\newblock Towards better text understanding and retrieval through kernel entity
  salience modeling.
\newblock In {\em ACM SIGIR'18}.

\bibitem[\protect\citeauthoryear{Xiong, Power, and
  Callan}{2017}]{xiong2017explicit}
Xiong, C.; Power, R.; and Callan, J.
\newblock 2017.
\newblock Explicit semantic ranking for academic search via knowledge graph
  embedding.
\newblock In {\em WWW'17}.

\end{thebibliography}

\end{document}